\documentclass{article}
\usepackage{graphicx}
\usepackage{latexsym}

\usepackage[utf8]{inputenc}
\usepackage{url}
\usepackage{amssymb,amsmath}
 \newcommand{\Bem}[1]{}

\usepackage{graphicx}
\usepackage[utf8]{inputenc}
\usepackage{amssymb}
\usepackage[boxed]{algorithm2e}

 \newcommand{\CLRL}{Combinatorial Laplacian Relative Lambda Method}
\newcommand{\CLSSAL}{Combinatorial Laplacian Sample Size Adjusted Lambda Method}
\newcommand{\CLSSAMXL}{Combinatorial Laplacian Sample Size Adjusted Maximum Lambda Method}
\newcommand{\NLL}{Normalized Laplacian Method}
\newcommand{\abbCLRL}{\texttt{CLRL}}
\newcommand{\abbCLSSAL}{\texttt{CLSSAL}}
\newcommand{\abbCLSSAMXL}{\texttt{CLMXL}}
\newcommand{\abbNLL}{\texttt{NLL}}

\newcommand{\figaddr}[1]{figs/#1}

\Bem{
\author{Mieczys{\l}aw A. K{\l}opotek\inst{1}
\orcidID{0000-0003-4685-7045}  \and
Bart{\l}omiej Starosta \inst{1}\orcidID{0000-0002-5554-4596} \and
S{\l}awomir T.  Wierzcho{\'n}\inst{1}\orcidID{2222--3333-4444-5555}}
\authorrunning{M. K{\l}opotek et al.}
\institute{Institute of Computer Science, Polish Academy of Sciences, ul. Jana Kazimierza 5, 01-248 Warsaw, Poland\\
\email{\{klopotek,barstar,stw\}@ipipan.waw.pl}, 
\url{http://www.ipipan.waw.pl} 
}
}

\allowdisplaybreaks

\def\abstract{
\typeout{Abstract}
 {\bf Abstract} 
}

\begin{document}
\title{Eigenvalue-based Incremental Spectral Clustering
}

\author{Mieczys{\l}aw A. K{\l}opotek
\and
Bart{\l}omiej Starosta 
\and
S{\l}awomir T.  Wierzcho{\'n}
\\
Institute of Computer Science, Polish Academy of Sciences,\\ ul. Jana Kazimierza 5, 01-248 Warsaw, Poland
}

\maketitle

%
%
\begin{abstract}  
Our previous experiments demonstrated that subsets collections of (short) documents (with several hundred entries) share a common normalized in some way eigenvalue spectrum of combinatorial Laplacian.
Based on this insight, we propose a method of incremental spectral  clustering. The method consists of the following steps: (1) split the data into manageable subsets, (2) cluster each of the subsets, (3) merge clusters from different subsets based on the eigenvalue spectrum similarity to form clusters of the entire set. 
This method can be  especially useful for clustering methods of complexity strongly increasing with the size of the data sample,like in case of typical spectral clustering. Experiments were performed showing that in fact the clustering and merging the subsets yields clusters close to clustering the entire dataset.   
\Bem{
\keywords{
Text mining \and 
Artificial Intelligence \and
Machine Learning \and 
Graph spectral clustering  \and Incremental clustering 
\and combinatorial Laplacian 
\and Eigenvalue spectrum analysis.}
}
\end{abstract}
%

\section{Introduction}

One of intensively \Bem{the heavily} developing clustering techniques is the Graph Spectral Analysis (GSA). It works best for objects whose mutual  relationships are described by a graph that connects them based on a similarity measure \Bem{It suits best for objects  whose  relations are best described by a graph linking them on the grounds of a  similarity measure)}~\cite{Mei:2022,Sevi:2022,STWMAKSpringer:2018}.

One important application is text documents clustering, where the similarity of documents can be expressed in a number of ways, e.g.  by the count of common words or in terms of more sophisticated descriptions (e.g. cosine similarity), see e.g.~\cite{Janani:2019}. 
In our experiments, we use the cosine between document vectors in the term vector space as the measure of document similarity. 
GSA applies  eigen-decomposition of the so-called graph Laplacians, derived from the similarity matrix. 

The original GSA suffers from the lack of   a method for assignment of new data items to the existing clusters. Hence a clustering from scratch or training of some external classification model is required.  Clustering from scratch may be hard for large data collections. Classification by the external model may cause cluster definition drift. Due to these issues,  several approaches were proposed to solve them, including ~\cite{Nie:2011,Bengio:2003,Alzate:2010,Shen:2020}.
This paper can be seen as a contribution to this type of research.
The mentioned approaches concentrate on transforming eigenvectors, while our method relies on eigenvalues only.

The algorithm proposed in this paper allows to perform the clustering in batches.  The algorithm has the following structure (details are given in section \ref{sec:ourmethod}):
\begin{itemize}
\item For each batch of documents, perform the traditional spectral clustering into the predefined number of clusters. 
\item
Compute the vector of combinatorial or normalized Laplacian eigenvalues of each cluster of each batch.   
\item Then, based on some dissimilarity criteria between the cluster spectra of different batches,  make a decision to combine the corresponding clusters of different batches.  
\item The matching of clusters is based on minimizing the difference between these vectors.
\end{itemize}
We investigated the following (dis)similarity criteria: 
\begin{itemize}
    \item normalize the spectra by dividing by the largest eigenvalue, then the dissimilarity is equal to an (approximate) integral between the class spectrum and the new data set spectrum (\CLRL, \abbCLRL)); see Fig~\ref{fig:MET_CLRL},
    \item normalize the spectra by dividing by the data{}set size (class or new data set), then the dissimilarity is equal to an (approximate) integral between the class spectrum and the new data set spectrum (\CLSSAL, \abbCLSSAL); see Fig~\ref{fig:MET_CLSSAL},
    \item normalize the spectra by dividing by the data{}set size (class or new data set), then the dissimilarity is equal to the absolute difference between largest eigenvalues  (\CLSSAMXL, \abbCLSSAMXL); see Fig~\ref{fig:MET_CLSSAL},
    \item compute not the combinatorial Laplacian but rather the Normalized Laplacian (which has always by definition the largest eigenvalue not greater than 2\footnote{the value 2 is attained for bipartite graphs}%
, then the dissimilarity is equal to an (approximate) integral between the class spectrum and the new data set spectrum (\NLL, \abbNLL); see Fig~\ref{fig:MET_NLL}. 
\end{itemize}

The dissimilarity measures mentioned above differ due to specific properties of GSA. \abbNLL{} is based on normalized Laplacian while the other three measures refer to combinatorial Laplacian. This has an effect on the shape of the respective spectrograms. 
Eigenvalues of normalized Laplacian are upper-bounded by the value of 2, whatever the sample size is. So if one has samples of different sizes from the same population, the value range is bounded and one needs only to adjust the indexes of eigenvalues to match the spectrograms of data from the same population. But the eigenvalues of combinatorial Laplacian can grow without any limit if the sample size increases. The approaches \abbCLRL, \abbCLSSAL{} and \abbCLSSAMXL{} handle the issue of matching spectrograms of data from the same population in different ways. It is necessary in all these cases to normalize the indexes of eigenvalues (into the range 0-1).   The approach  \abbCLRL{} normalizes the eigenvalues by dividing by the largest eigenvalue, while  \abbCLSSAL{} 
divides them by the sample size. The effects of both on the spectrogram would be the same for samples from the population, but the shapes of different population spectrograms will differ in different ways (e.g. in  \abbCLRL{} the spectrograms will meet at both ends, while in \abbCLSSAL{} they will not). 
divides them by the sample size. The effects of both on the spectrogram would be the same for samples from the population, but the shapes of different population spectrograms will differ in different ways (e.g. in  \abbCLRL{} the spectrograms will meet at both ends, while in \abbCLSSAL{} they will not).  \abbCLRL{} is more susceptible to noise at the largest eigenvalue than  \abbCLSSAL{}. 
\abbCLSSAMXL{} transforms the spectrogram in the same way as \abbCLSSAL{}, but instead of using an integral to assess the differences between populations it take the larges eigenvalue after normalization. For justifications of the used properties see \cite{PlosOne:2023}.

Our algorithm is proposed in section \ref{sec:ourmethod}.

The experimental study of the effectiveness of our method is presented in section \ref{sec:experiments} and the conclusions are described in section \ref{sec:conclusion}.  
Let us first provide with an overview of concepts behind spectral clustering methods in section~\ref{sec:previous}. 

%
%
\section{Previous Work}\label{sec:previous}

One observes growing interest in graph spectral clustering and classification methods. While they have interesting properties with respect to spatial form of clusters and classes \cite{Lux07}, they face the problem of inability to operate incrementally~\cite{Nie:2011,Bengio:2003,Alzate:2010,Shen:2020}. Let us briefly explain the reasons for this problem. 

The traditional way to perform graph spectral clustering is based on   relaxation  of ratio cut (RCut) and normalized cut (NCut) graph clustering methods.  $k$-means algorithm is applied to the rows of the matrix, the columns of which are eigenvectors associated with the $k$ lowest eigenvalues of the corresponding graph Laplacian \cite{Lux07}. 

Formally,  consider  a similarity matrix $S$  between pairs of items (e.g. documents). One can imagine a weighted graph $G$ linking the items with weights represented by $S$.  A(n unnormalized) or combinatorial Laplacian $L$ of the matrix $S$  is defined as 
\begin{equation}\label{eq:combLapDef} L=T-S, \end{equation}
where $T$ is the diagonal matrix with $t_{jj}=\sum_{k=1}^ns_{jk}$ for each $j \in [n]$. 
A normalized Laplacian $\mathcal{L}$ of the graph represented by~$S$ is defined as 
\begin{equation}\label{eq:normLapDef}\mathcal{L}=T^{-1/2}L T^{-1/2}= I -T^{-1/2}S T^{-1/2}. \end{equation}%
Recall that the RCut criterion means finding the partition matrix $P_{RCut} \in \mathbb{R}^{n \times k}$ that minimizes the formula $H' L H$ over the set of all partition matrices $H \in \mathbb{R}^{n\times k}$. This minimization problem turns out to be NP-hard. This is the reason for   relaxing it by assuming that $H$ is a column orthogonal matrix. Then  the solution is simple: the columns of $P_{RCut}$ are eigenvectors of $L$ corresponding to $k$ smallest eigenvalues of~$L$. Similarly, the columns of matrix $P_{NCut}$, representing NCut criterion, are eigenvectors of $\mathcal{L}$ corresponding to $k$ smallest eigenvalues of~$\mathcal{L}$. For an explanation and further details see e.g.~\cite{Lux07} or~\cite{STWMAKSpringer:2018}. 

Various modifications are applicable, including  
(1) usage of the top eigenvalue eigenvectors of the matrix $D^{-1/2}S D^{-1/2}$ instead of the lowest ones\cite{Kamvar2003,Suganthi:2018}, (2) normalization of the rows of the aforementioned eigenvector sub-matrix   to unit length prior to $k$-means clustering, (3) making use of more than $k$ eigenvectors to cluster into $k$ clusters,~\cite{RV11}, (4)  application of a supervised learning method, instead of clustering, preferentially on a subset of the rows of the aforementioned sub-matrix, followed by employing the learned classifier to the remaining rows. 

There exists research also on semi-supervised spectral clustering, like the semi-supervised sentiment classification of  
Li and Hao~\cite{LH12}  or 
semi-supervised spectral
detection of population stratification by Liu, Shen, and Pan  in~\cite{LSP13}.

The growing interest in spectral clustering results from the   ability
to deal  with nonlinearly separable datasets  But regrettably it suffers from a critical limitation induced by   its huge time and space complexity. This handicap severely restricts applicability to  large-scale problems.
The fact that all these methods rely on computation of eigenvectors and that eigenvectors do not exhibit the property of eigenvectors of bigger matrices being derivable from smaller matrices, there exists a problem with new coming data that they enforce computations of the eigen-vectors from scratch.  
This limitation prompted researchers to develop methods that can help to overcome this limitation. 
One strategy strategy relies on sparsifying the
affinity $S$ matrix and solve the eigen-decomposition problem
by sparse eigen-solvers \cite{Lux07}. 
Another strategy is to construct   sub-matrices. E.g., the method of  
 Nystr{\"o}m, as applied by \cite{Chen:2011}, randomly selects p representatives
from the original dataset and builds an N × p affinity sub-matrix. 
\cite{Cai:2015} improved this method by  proposing so-called  landmark based spectral clustering (LSC) method,
which performs k-means on the dataset to get p cluster
centers as the p representatives. 
Both approaches seem to suffer from the bottleneck of the number sub-matrices to be sampled. 
\cite{Huang:2020} proposed two algorithms,   ultra-scalable spectral clustering (U-SPEC) and ultra-scalable ensemble
clustering (U-SENC). U-SPEC relies on  fast approximation method for K-nearest
representatives used in   construction of a sparse affinity sub-matrix.
U-SENC integrates multiple   U-SPEC
into an ensemble clustering framework. 
These algorithms were further refined in \cite{Sitnik:2022clustering} by exploring the approximate explicit feature map (aEFM) transform of low-dimensional data into a low-dimensional subspace in Hilbert space.
 Still another approach relies on divide-and-conquer paradigm applied to the landmark-based methodology \cite{Li:2022}. 
 The path is followed in \cite{Hong:2023} where probability density estimation drives the landmark approach. 
 The idea of dealing with processing complexity via ensemble clustering  is followed up in 
 \cite{Li:2021lsec}.

The paper  \cite{PlosOne:2023} proposes a completely different approach to the problem of this eigenvector discontinuity. 
Instead of relying on eigenvectors, it turns to sole usage of eigenvalues. The paper 
investigates batch type classification problem. Given a collection of documents, labelled with classes, consider a new batch of documents which is known to belong to a single class, but it is unknown to which. It turns out that comparison of the spectrum of the combinatorial Laplacian of the unlabeled batch with those of labelled batches can identify the appropriate batch with reasonable probability.

\section{Our Method}\label{sec:ourmethod}

The theoretical background to our assumptions is outlined in the mentioned paper \cite{PlosOne:2023}. 
We do not follow the Nystr{\"o}m  paradigm of operating in the embedding space of $L$ matrices. Instead we look at the eigenvalue spectra. 

The essence of the approach to clustering is to split the document set $D$, that may be too large to handle, into smaller batches (data portions) $D_i$. Each of them should be clustered with a spectral clustering method. Subsequently, we identify corresponding clusters of each document batch via comparing eigenvalue spectra. 
In this way we produce the clusters of the big dataset $D$.

The Algorithm \ref{alg:ourmethodframework} presents in a compact way the described method bundle. 
The functions called in the sub-algorithm \ref{alg:ourclustermatchingmethod}, that is
$L()$, $spectrum()$, $specfun()$, $spectdist()$
are described below.

\begin{algorithm}
\KwData{$D$ - a (large) set of documents, to be processed in batches\\
$k$ - the number of clusters to be obtained  }
\KwResult{ $\Gamma$ - the clustering of $D$ into $k$ clusters }
Split randomly $D$ into (small) subsets $\{D_0,\dots,D_m\}$\;
For each $D_i$ compute its spectral clustering $\Gamma_i$ into $k$ clusters\;
For each cluster $C_{i,j}\in \Gamma_i$ compute the similarity matrix $S_{i,j}$
$\Gamma:=\Gamma_0$ - initial  clusters ($\Gamma_0$ is the $D_0$ spectral custering)  \;
\For{$i\leftarrow 1$ \KwTo $m$} 
{
  \For{$j\leftarrow 1$ \KwTo $k$} 
  { call Algorithm \ref{alg:ourclustermatchingmethod} setting: 
    $S:=S_{ij}$; 
    $\mathfrak{S}:=\{S_{0,1}, \dots,S_{0k} \} $\;
    $c$ be the identifier returned by it\;
    Update $C_{c}\in \Gamma$  with $C_{c}\cup C_{j,c}$\;
  }
}
\caption{The eigenvalue based clustering  algorithm}\label{alg:ourmethodframework}
\end{algorithm}

\begin{algorithm}
\KwData{$S$ - similarity matrix of the new cluster of documents\\
$\mathfrak{S}$ - set of similarity matrices of the clusters of documents to match with }
\KwResult{$c$ - the assigned cluster of documents  }
$L:=L(S)$ - Compute Laplacian\;
$\mathfrak{L}:=L(\mathfrak{S})$ - Compute Laplacians\;
$E:=spectrum(L)$ - Compute Laplacian eigenvalues\;
$\mathfrak{E}:=spectrum(\mathfrak{L})$ - Compute Laplacian eigenvalue for each Laplacian from $\mathfrak{L}$\;
$F:=specfun(E)$ - transform a spectrum into a function\;
$\mathfrak{F}:=specfun(\mathfrak{E})$ - transform spectra into  functions\;
$K\leftarrow $ number of clusters in $\mathfrak{S}$\;
$c\leftarrow -1$\;
$mndist\leftarrow \infty$\;
\For{$j\leftarrow 1$ \KwTo $K$} 
{
 $distance\leftarrow spectdist(F,\mathfrak{F}_j)$ \;
\eIf{$distance <mndist$}{
$c\leftarrow j$ \;
$mndist\leftarrow distance$ \;
}{
do nothing\;
}
}
\caption{The eigenvalue based class assignment   algorithm}\label{alg:ourclustermatchingmethod}
\end{algorithm}

A drawback of this approach is that each cluster to be discovered  must be a homogeneous group and must be distributed proportionally over various batches.
By a homogeneous group we understand a population i which each sample has the same (exactly speaking very similar) spectrogram (after normalization by a given method). In fact, our experiments reported in \cite{PlosOne:2023} demonstrate that this is the case for various datasets. Homogenicity does not mean that each batch must be of the same size. Rather the share of the group in each batch should be the same. If the shares of groups in different batches differ, then the spectral clustering algorithms would havbe in general a problem because their underlying algorithm, $k$-means, does not "like" clusters that differ too much in size and shape. This is not a flaw of our approach, but rather a general problem of GSA. 
Nonetheless  there exist practical applications where homogenous groups occur proportionally in batches.   
One example is the task of clustering products handled by big sales companies. The number of consumer products in large chains of hypermarkets may amount to hundreds of thousands and new ones occur in bundles every week. The suppliers do not care about the groups of products the chain has created so it is the job of chain employees to cluster the products based on their descriptions. While large computers may handle spectral clustering in hundreds of thousands of dimensions, accessibility of such machines may be not common enough, so that approaches to lower the scale need to be sought.  

 In the Algorithm \ref{alg:ourclustermatchingmethod}, 
 being a subroutine of our main Algorithm \ref{alg:ourmethodframework}, the following functions are used:
\begin{itemize}
    \item 
$spectdist(F_1,F_2)$ function is the area between the two functions $F_1,F_2$ being its arguments for the function domains [0,1], $\int_0^1 |F_1(x)-F_2(x)|dx$, except for
\abbCLSSAMXL, where $|F_1(0)-F_2(0)|$ is returned. 
\item 
 The function $L(S)$ applied to the similarity matrix $S$ is computed as $D(S)-S$ except for \NLL (\abbNLL), where $D(S)^{-1/2}Z(D(S)-S)D(S)^{-1/2}Z$ is the result.
 \item 
The function $spectrum(L)$ applied to Laplacian $L$ returns a vector of eigenvalues of $L$ in non-decreasing order. 
\item 
The function $specfun(E)$ applied to the spectrum $E$ of a Laplacian returns a function $F(x)$ defined in the domain $x \in [0,1]$ with properties depending on the type of cluster-matching method.
\begin{itemize}
    \item for \abbCLRL: 
    $$F\left(\frac{n-i}{n-1}\right)=\frac{\lambda_i}{\lambda_n}  $$
    \item for \abbCLSSAL\/ and \abbCLSSAMXL: 
    $$F\left(\frac{n-i}{n-1}\right)=\frac{\lambda_i}{n}  $$
\item for \abbNLL:
    $$F\left(\frac{n-i}{n-1}\right)=\lambda_i  $$
\end{itemize}
and otherwise for any $x\in \left[\frac{n-(i+1)}{n-1},\frac{n-i}{n-1}\right]$
\[
\begin{split}
    F(x)&=F\left(\frac{n-(i+1)}{n-1}\right)\cdot \left(x-\frac{n-(i+1)}{n-1}\right)\\
    &+F\left(\frac{n-i}{n-1}\right)\cdot \left(\frac{n-i}{n-1}-x\right).
\end{split}\]
$n$ is the number of elements in the spectrogram $E$; the spectrogram is the sequence of eigenvalues ordered decreasingly, with their index $i$ running from 1 to $n$.  
\end{itemize}

Note that our approach to distance computation between spectra (function $spectdist$) bears some resemblance to Dynamic Time Warping (DTW, \cite{rabiner_juang93}) distance, but the difference is that we apply a linear transformation to the index axis of the spectrogram, while DTW promotes non-linear transformations.

\section{Dataset}

For our experiments, we used tweets provided by Twitter
\Bem{We used for our experiments Twitter data as made available by Twitter} (a random sample of about 1\% of English tweets) collected for the time period from mid September 2019 till end of May 2022.

We restricted our investigation \Bem{for this study} to tweets having only one hashtag at the end of text with at least 10 words, whereby we restrict ourselves to the hashtags: \#bbnaija, \#blacklivesmatter and    
\#puredoctrinesofchrist. 
This dataset will be referred to as \emph{TWT.EN}. 
A copy can be found in \texttt{Supplementary File}.

\section{Experiments}\label{sec:experiments} 

We want to demonstrate via the experiments that our algorithm correctly matches the clusters stemming from different data portions. 

The ideal case for such a demonstration wold be that the base clustering algorithm splits the dataset along known  labels (coming from an external labeling) and then our method matches clusters from different data portions combining the clusters with the same external label. This external label is of course not known to the algorithm. 

The only external labels available for our tweets are the hashtag. So the ideal situation would be if an  algorithm may split the TWT.EN data in agreement with hashtags. 

We assume in our first stage of experiments (subsections \ref{sec:stage1:matching} and \ref{sec:stage1:stability}) that in fact such an ideal algorithm exists and has split each portion exactly in agreement with the hashtag labeling. 
Then we  check if our algorithm can correctly match "clusters" stemming from different batches (data portions). 

In the second stage  (subsections \ref{sec:stage2:matching} and \ref{sec:stage2:stability}),  we exploit a real spectral clustering algorithm, approximating the split by hashtag.  

\subsection{Differentiation of hashtags by Laplacian spectrum}\label{sec:stage1:matching}
In order to check the differentiation of hashtags by Laplacian eigenvalue spectrum, the dataset TWT.EN was divided randomly into three subsets (data portions) of approximately same size. 
The distribution of the number of documents in data portions for each hashtag is shown in Table~\ref{tab:portionhashtags}. 

\begin{table}
\center%
\begin{tabular}{|l|r|r|r||r|}
\hline Data portion& \#bbnaija& \#blacklivesmatter& \#puredoctrines & total\\
& & &ofchrist &\\
\hline all 
                  &1857 &                  2051     &              1295 &
         5203         \\

\hline Portion  1  
               &    649      &              677      &              409 
&  1735 \\
\hline Portion  2  
                &   610       &             686       &             439 
& 1735 \\
\hline Portion  3  
                &   598         &           688        &            447 
& 1733 \\
\hline
\end{tabular}
\caption{Hashtag distribution over data portions}
\label{tab:portionhashtags}

\end{table} 

\Bem{For each subset (data portion) and each hashtag, the combinatorial and normalized  Laplacian was computed and its eigenvalue spectrum. }
For each subset (data portion) and each hashtag, the combinatorial and normalized Laplacian and its spectrum of eigenvalues were calculated.
The results, in normalized form, suitable for respective methods, are shown in Figs~\ref{fig:MET_CLRL},~\ref{fig:MET_CLSSAL} and~\ref{fig:MET_NLL}. 
The figures represent the aforementioned functions $specfun()$ for \abbCLRL{} (Fig. \ref{fig:MET_CLRL}), \abbCLSSAL{}
(Fig.\ref{fig:MET_CLSSAL}) and \abbNLL{} (Fig.\ref{fig:MET_NLL}) for each of the eigenvalue spectrum of a given hashtag of a given data portion. Lines related to the same hashtag have the same color. To improve visibility, the hashtag names were replaced in the figures \Bem{by coding explained in Table \ref{tab:dic:classesTWT.EN}} by the coding \texttt{gr1} $\leftarrow$ \texttt{\#bbnaija}, \texttt{gr2} $\leftarrow$ \texttt{\#blacklivesmatter}, and \texttt{gr3} $\leftarrow$ \texttt{\#puredoctrinesofchris}. 
As you can see, the method \abbCLSSAL{} is characterized by the best separation of the spectrograms of different hashtags. The hashtag \#puredoctrinesofchrist seems to be best separated from the other ones.

\subsection{Stability of hashtag spectra over various samples} \label{sec:stage1:stability}

In order to verify the usability of various cluster matching methods (\abbCLRL, \abbCLSSAL, \abbNLL), the stability of hashtag eigenvalue spectra over various samples was investigated.  

First, based on the data portion 1, our cluster-matching algorithm was "trained" (the spectra of hashtags were acquired, that is for each hashtag in the data portion 1 the spectrogram of Laplacian was computed for each data subset marked with this  hashtag). 
The Algorithm~\ref{alg:ourmethodframework} was applied then to an artificial series of 100 data portions created as random {sub}samples of data portions 2 and 3. 

The correctness of data portion assignment to hashtags is shown in Tables~\ref{tab:classesCLRLM.0.5.TWT.EN}-\ref{tab:classesCLSSAMXLM.0.5.TWT.EN} 
for the respective methods.

\begin{figure}
\begin{center}
\includegraphics[width=0.5\textwidth]{\figaddr{NTTWT.ENAllPorclassesComb.png}} 
 \end{center}
\caption{Spectral normalization in the \CLRL{}  method. The TWT.EN dataset }\label{fig:MET_CLRL}
\end{figure}

\Bem{
\begin{table} %
\center%
\begin{tabular}{|l|l|}
\hline 
gr1 &  \#bbnaija\\
\hline 
gr2 &  \#blacklivesmatter\\
\hline 
gr3 &  \#puredoctrinesofchrist\\
\hline

\end{tabular}
\caption{A dictionary for  TWT.EN  datset} 
\label{tab:dic:classesTWT.EN}
\end{table}
}

\begin{table} 
\center%
\begin{tabular}{|r|r|r|r|}
\hline TRUE/PRED& gr1& gr2& gr3\\
\hline  gr1& 56& 44& 0\\
\hline  gr2& 26& 74& 0\\
\hline  gr3& 0& 91& 9\\
\hline
\end{tabular}
\caption{Classification experiment for the dataset  TWT.EN for classes using Combinatorial Laplacian Relative Lambda Method}
\label{tab:classesCLRLM.0.5.TWT.EN}

\end{table} 

\begin{table} 
\center%
\begin{tabular}{|r|r|r|r|}
\hline TRUE/PRED& gr1& gr2& gr3\\
\hline  gr1& 0& 0& 100\\
\hline  gr2& 0& 0& 100\\
\hline  gr3& 0& 0& 100\\
\hline
\end{tabular}
\caption{Classification experiment for the dataset  TWT.EN for classes using Normalized Laplacian Method}
\label{tab:classesNLM.0.5.TWT.EN}

\end{table} 


 \begin{figure}
\begin{center}
\includegraphics[width=0.5\textwidth]{\figaddr{NTTWT.ENAllPorclassesComb_ssa.png}} 
 \end{center}
\caption{Spectral normalization in the \CLSSAL{} method and \CLSSAMXL{} method. The TWT.EN dataset. }\label{fig:MET_CLSSAL}
\end{figure}

\begin{table} 
\center%
\begin{tabular}{|r|r|r|r|}
\hline TRUE/PRED& gr1& gr2& gr3\\
\hline  gr1& 100& 0& 0\\
\hline  gr2& 0& 100& 0\\
\hline  gr3& 0& 0& 100\\
\hline
\end{tabular}
\caption{Classification experiment for the dataset  TWT.EN for classes using Combinatorial Laplacian Set Size Adjusted Lambda  Method}
\label{tab:classesCLSSALM.0.5.TWT.EN}

\end{table} 

\begin{table} 
\center%
\begin{tabular}{|r|r|r|r|}
\hline TRUE/PRED& gr1& gr2& gr3\\
\hline  gr1& 73& 27& 0\\
\hline  gr2& 14& 86& 0\\
\hline  gr3& 0& 0& 100\\
\hline
\end{tabular}
\caption{Classification experiment for the dataset  TWT.EN for classes using Combinatorial Laplacian SSA Maximal Lambda  Method}
\label{tab:classesCLSSAMXLM.0.5.TWT.EN}

\end{table} 

\begin{figure}
\begin{center}
\includegraphics[width=0.5\textwidth]{\figaddr{NTTWT.ENAllPorclassesNorm.png}} 
 \end{center}
\caption{Spectral normalization in the \NLL{}. The TWT.EN dataset.}\label{fig:MET_NLL}
\end{figure}

\begin{table} %
\center%
\begin{tabular}{|l|r|r|}
\hline Method & Error \% & F1 value \\
\hline CLRL & 53.67   &  41.98  \\
CLSSAL & 0   &  100  \\
CLMXL & 13.67   &  86.27  \\
NLL & 66.67   &  16.67  \\
\hline
\end{tabular}
\caption{Errors and F1 values for  TWT.EN  datset} 
\label{tab:errF1:TWT.EN}

\end{table} 

As visible from Table~\ref{tab:classesCLSSALM.0.5.TWT.EN}, the method \abbCLSSAL{} provides with best results. \abbCLSSAMXL{} is second best (Table \ref{tab:classesCLSSAMXLM.0.5.TWT.EN}). See the error and F1 Table~\ref{tab:errF1:TWT.EN}.

This allows us to conclude that if a clustering method would approximate well the hashtag allocations for these hashtags, then incremental clustering would be possible. 

\subsection{Differentiation of clusters by Laplacian spectrum}\label{sec:stage2:matching}

As a next step,  each data portion was clustered by  
Normalized spectral clustering method with  unit length rows and one additional dimension (that is by a real-world spectral clustering algorithm). 

The result of these clustering processes are visible 
in Tables~\ref{tab:clustersportion1},  \ref{tab:clustersportion2} and  \ref{tab:clustersportion3} for data portions 1,2, and 3 respectively.

\begin{table}
\center%
\begin{tabular}{|r|r|r|r|}
\hline TRUE/PRED& pseudo-1& pseudo-2& pseudo-3\\
\hline  \#bbnaija& 429& 219& 1\\
\hline  \#blacklivesmatter& 247& 430& 0\\
\hline  \#puredoctrinesofchrist& 38& 86& 285\\
\hline
\end{tabular}
\caption{Result of clustering data portion no 1}
\label{tab:clustersportion1}

\end{table} 

\begin{table} 
\center%
\begin{tabular}{|r|r|r|r|}
\hline TRUE/PRED& pseudo-1& pseudo-2& pseudo-3\\
\hline  \#bbnaija& 410& 200& 0\\
\hline  \#blacklivesmatter& 208& 477& 1\\
\hline  \#puredoctrinesofchrist& 94& 54& 291\\
\hline
\end{tabular}
\caption{Result of clustering data portion no 2}
\label{tab:clustersportion2}

\end{table} 

\begin{table} 
\center%
\begin{tabular}{|r|r|r|r|}
\hline TRUE/PRED& pseudo-1& pseudo-2& pseudo-3\\
\hline  \#bbnaija& 448& 150& 0\\
\hline  \#blacklivesmatter& 277& 411& 0\\
\hline  \#puredoctrinesofchrist& 21& 132& 294\\
\hline
\end{tabular}
\caption{Result of clustering data portion no 3}
\label{tab:clustersportion3}

\end{table} 

We assigned cluster labels in such a way that in each clustering the cluster with the same highest share of a given hashtag gets the same cluster label. These cluster labels were of course invisible to the cluster-matching algorithm. 

For each subset (data portion) and each cluster label, the combinatorial and normalized  Laplacians  and their eigenvalue spectra were computed.  
As previously, the results, in normalized form, suitable for respective methods, are shown in Figs~\ref{fig:MET_CLRL_clusters},~\ref{fig:MET_CLSSAL_clusters} and~\ref{fig:MET_NLL_clusters}. 

Lines related to the same cluster label have the same color.  

One can see that again the method \abbCLSSAL{} of data normalization is a clear winner, though first two clusters are not separated well.

\subsection{Stability of cluster spectra over various samples} \label{sec:stage2:stability}

To investigate the ability of our method to match clusters from various data portions appropriately, we again   trained our cluster-matching algorithm  based on the data portion~1,  
The Algorithm \ref{alg:ourmethodframework} was applied then to an artificial series of 100 data portions created as random {sub}samples of data portions~2 and~3. 

The correctness of data portion assignment to clusters is shown in Tables 
\ref{tab:clustersCLRLM.0.5.TWT.EN}-\ref{tab:clustersCLSSAMXLM.0.5.TWT.EN}  
for the respective methods.

\begin{figure}
\begin{center}
\includegraphics[width=0.5\textwidth]{\figaddr{NTTWT.ENAllPorclustersComb.png}} 
 \end{center}
\caption{Spectral normalization in the \CLRL{}  method. The TWT.EN dataset }\label{fig:MET_CLRL_clusters}
\end{figure}

\begin{table} 
\center%
\begin{tabular}{|r|r|r|r|}
\hline TRUE/PRED& pseudo-1& pseudo-2& pseudo-3\\
\hline  pseudo-1& 40& 57& 3\\
\hline  pseudo-2& 51& 41& 8\\
\hline  pseudo-3& 0& 0& 100\\
\hline
\end{tabular}
\caption{Classification experiment for the dataset  TWT.EN for clusters using Combinatorial Laplacian Relative Lambda Method}
\label{tab:clustersCLRLM.0.5.TWT.EN}

\end{table} 

\begin{table} 
\center%
\begin{tabular}{|r|r|r|r|}
\hline TRUE/PRED& pseudo-1& pseudo-2& pseudo-3\\
\hline  pseudo-1& 0& 0& 100\\
\hline  pseudo-2& 0& 0& 100\\
\hline  pseudo-3& 0& 0& 100\\
\hline
\end{tabular}
\caption{Classification experiment for the dataset  TWT.EN for clusters using Normalized Laplacian Method}
\label{tab:clustersNLM.0.5.TWT.EN}

\end{table} 

\begin{figure}
\begin{center}
\includegraphics[width=0.5\textwidth]{\figaddr{NTTWT.ENAllPorclustersNorm.png}} 
 \end{center}
\caption{Spectral normalization in the \NLL{}. The TWT.EN dataset.}\label{fig:MET_NLL_clusters}
\end{figure}

\begin{table} 
\center%
\begin{tabular}{|r|r|r|r|}
\hline TRUE/PRED& pseudo-1& pseudo-2& pseudo-3\\
\hline  pseudo-1& 36& 64& 0\\
\hline  pseudo-2& 84& 16& 0\\
\hline  pseudo-3& 0& 0& 100\\
\hline
\end{tabular}
\caption{Classification experiment for the dataset  TWT.EN for clusters using Combinatorial Laplacian Set Size Adjusted Lambda  Method}
\label{tab:clustersCLSSALM.0.5.TWT.EN}

\end{table} 

\begin{figure}
\begin{center}
\includegraphics[width=0.5\textwidth]{\figaddr{NTTWT.ENAllPorclustersComb_ssa.png}} 
 \end{center}
\caption{Spectral normalization in the \CLSSAL{} method and \CLSSAMXL{} method. The TWT.EN dataset. }\label{fig:MET_CLSSAL_clusters}
\end{figure}

\begin{table} 
\center%
\begin{tabular}{|r|r|r|r|}
\hline TRUE/PRED& pseudo-1& pseudo-2& pseudo-3\\
\hline  pseudo-1& 37& 63& 0\\
\hline  pseudo-2& 15& 85& 0\\
\hline  pseudo-3& 0& 0& 100\\
\hline
\end{tabular}
\caption{Classification experiment for the dataset  TWT.EN for clusters using Combinatorial Laplacian SSA Maximal Lambda  Method}
\label{tab:clustersCLSSAMXLM.0.5.TWT.EN}

\end{table} 

\begin{table} %
\center%
\begin{tabular}{|l|r|r|}
\hline Method & Error \% & F1 value \\
\hline CLRL & 39.67   &  59.36  \\
CLSSAL & 49.33   &  50.17  \\
CLMXL & 26   &  72.41  \\
NLL & 66.67   &  16.67  \\
\hline
\end{tabular}
\caption{Errors and F1 values for  TWT.EN  datset} 
\label{tab:errF1:TWT.EN:cluster}

\end{table} 

As one can see, the method \abbCLSSAL{} is the best one, but due to overlapping nature of spectrograms of clusters pseudo-1 and pseudo-2, they are not as well matched as the cluster pseudo-3. See also the error and F1 Table~\ref{tab:errF1:TWT.EN:cluster}.


\section{Conclusions}\label{sec:conclusion}  
%

Our research  shows that Twitter tweets related to the same hashtag are "similar" in "style" in any {sub}sample (that is their combinatorial Laplacian spectrum). This carries over to clusters obtained by clustering algorithms discovering the hashtags. So instead of clustering of the entire set, we could propose in this paper to cluster subsets and then recover the total cluster and then match the clusters from subsets via the Laplacian spectrum. 

It seems to be an interesting feature of collections of (short) documents that their subsets share a common normalized in some way eigenvalue spectrum. If clusters obtained from a given method yield spectra of significantly different characteristics, then the spectral analysis may be exploited for the purpose of splitting the clustering process to smaller portions of data and then matching the obtained {sub}clusters via the outlined \CLSSAL{} method. This can be especially useful for clustering methods of complexity strongly increasing with the size of the data sample. 

Further research would be directed towards understanding the mechanism relating the literary style of short documents collections on  a particular topic to the eigenvalue spectrogram and the reasons why some spectrograms are more or less similar for various topical collections.

\bibliographystyle{plain}

 
\end{document}